# Building upon Fast Multipole Methods to detect and model organizations

Pierrick Tranouez[1,2] and Antoine Dutot[1]

[1] LITIS
Université du Havre
UFR Sciences et Techniques
25 rue Ph. Lebon - BP 540
76058 Le Havre Cedex – France
{Pierrick.Tranouez, Antoine.Dutot}@litislab.eu

[2] MTG - UMR IDEES
Université de Rouen
IRED
7 rue Thomas Becket
76821 Mont Saint Aignan Cedex – France

**KEYWORDS**

Organization detection, structure management, multiscale, N body systems, hierarchical tree code approximations

**ABSTRACT**

Many models in natural and social sciences are comprised of sets of interacting entities whose intensity of interaction decreases with distance. This often leads to structures of interest in these models composed of dense packs of entities. Fast Multipole Methods are a family of methods developed to help with the calculation of a number of computable models such as described above. We propose a method that builds upon FMM to detect and model the dense structures of these systems.

**INTRODUCTION**

We study in this paper dynamic systems composed of many interacting entities. We are interested in their modeling, simulation, in the way they evolve and organize themselves.

Among these theoretical systems, a class of them is used as models for different natural science systems. In this class, systems verify the following properties:

1. Entities follow an analytical model
2. Each entity interacts with all the others.
3. The strength of the interactions decreases with the distance separating the entities implied in the relationship

It is to this class that N-Body problems in physics belong to for example.

Property 2 makes computing a simulation through the iterative calculations of the model in 1 costly, in $O(n^2)$ for $n$ entities. There are mathematical techniques to accelerate this computing, such as Fast Fourier Transform or Fast Multipole Methods. We will deal in this paper with hierarchical methods such as FMM.

Furthermore, systems of the class described above have also some remarkable properties in their organizations. As interactions intensity decrease with the distance and these interactions are often of an attractive nature, aggregate or clusters often emerge of their simulation, and play important parts as structures and organizations of the simulation.

We give in this article a method that detects and manages such organizations in this class of models, in connection with an FMM simulation (or hierarchical tree code approximations in general).

**KERNELS AND FAST MULTIPOLE METHODS**

The Introduction may give the impression that our organization-handling scheme works in only very restricted cases, but systems verifying these properties often appear in scientific literature [Beatson and Greengard].

**Physics examples**

Provided N bodies of mass $m_i$ a description of the gravitational field in $x_j$ is Newton law [Newton 1686]:

$$g(x_j) = \sum_{\substack{i=1 \\ i \neq j}}^{N} \mathcal{G} m_i \frac{x_i - x_j}{\|x_j - x_i\|^3}$$

Where $\mathcal{G}$ is the gravitational constant.

Similarly for N bodies of charge q, Coulomb law [Coulomb 1785] describes the electrostatic field in $x_j$:

$$E(x_j) = \sum_{\substack{i=1 \\ i \neq j}}^{N} \mathcal{K} m_i \frac{x_i - x_j}{\|x_j - x_i\|^3}$$

Where $\mathcal{K}$ is the electrostatic constant.

In both of these, each body interacts with all the other, with an intensity that is inversely linear to the square of the distance. It is the same for Biot-Savart law, relative to both magnetism (under) and fluid dynamics [Leonard 1980]:

$$B(x) = \frac{1}{c} \int J(y) \frac{x-y}{|x-y|^3} dy$$

Where $J(y)$ is the current density in $y$.

The same can be written for diffusion, of heat for instance.

For all of these, structures of close entities play an important part. In gravitation, on different time scale they are the galaxies, solar systems, stars, planets. In magnetism, hydrodynamics and aerodynamics, congruence leads to vortexes or coils of different sizes, ranging from turbulence to the Gulf Stream or the red spot of Jupiter. In the Sun, congruence of plasma matter lead to sunspots and solar flare, and combined with fluids and gravity to coronal mass ejections.

**Life sciences examples**

Many living systems can be described the same way [Simon 1996] [Frontier and Pichot-Viale 1998], with hierarchical spatial organizations. One of these models is the boids [Reynolds 1987]. We will apply our method to this model, so we will describe it in some depth.

Boids are entities moving in a 2D or 3D space. They are ruled by quite simple rules, *Cohesion, Separation* and *Alignment*. Cohesion describes how boids try to fly in the direction of surrounding boids. Separation describes how boids are repulsed by other boids if these are really close. Alignment describes how boids try to fly in the same direction as the boids that environs them. A more formal description is:

Let j be a boid situated in the point $x_j$ of an affine space. Let $\eta(j)$ be its neighboring space, ie the set of boids not further from j than a certain limit at a given time t, excluding j itself.
Let $v_t^j$ be the speed of $j$ at time $t$. All the $x_k$ below should be written $x_t^k$ but are conventionally simplified to easy up the reading.

We define:

$$c_{t+1}^j = \sum_{i \in \eta(j)} \frac{x_i}{\|x_i - x_j\|^2} - x_j$$

$$s_{t+1}^j = \sum_{i \in \eta(j)} \frac{(x_j - x_i)}{\|x_i - x_j\|^3}$$

$$a_{t+1}^j = \sum_{i \in \eta(j)} \frac{v_t^i}{card(\eta(j)) \|x_i - x_j\|^2}$$

Which we combine:

$$v_{t+1}^j = \alpha v_t^j + \beta c_{t+1}^j + \gamma s_{t+1}^j + \delta a_{t+1}^j$$

Where $\alpha, \beta, \gamma, \delta$ are coefficients used to balance the different components of the speed.

This is for *one* species of boids. If *different* species of boids are considered, another stronger inter-species Separation is furthermore implemented to represent a tendancy to flock to their species and flee others [Dutot 2005]. Each species member speed is only influenced by its co-members as far as Alignment, Cohesion and intra species Separation are concerned. Each species can also have their own coefficients $\alpha, \beta, \gamma$ and $\delta$.

**Hierarchical methods to accelerate the computing**

The computing of the relation above is of a complexity of the order of the square of the number of entities/bodies involved. This can be costly as the size of the simulation increases.

An often used technique to improve this is is to recursively divide the simulation space in a tree. The root of the tree encompasses the whole space. Its children are regular subdivision of the space, for example in 4 equal surfaces. The process is repeated recursively. We use an adaptive subdivision, which means regions of space containing many entities are more divided that sparser ones. The entities are associated to the leaves of the tree, which is adapted at each simulation step.

The degenerated kernel relations described above are then computed. For that, the contribution of close entities is added exactly, while the contribution of further entities is averaged at the level of the leaf it belongs to. The exact economy is difficult to compute precisely as it depends on the structure of the tree, but it can be brought back to an average of $O(NlogN)$ or even $O(N)$ [Beatson and Greengard] for $N$ entities in the simulation. Furthermore, the loss of precision can be bound.

We originally used FMM in [Tranouez 2005] in a fluid flow simulation [Tranouez et al. 2005]. We discovered at the time that the tree used for the poles visually summarized the organizations appearing during the simulation (groups of vortexes of different shapes and sizes in this case). We therefore created an algorithm that formalized this visual discovery. We will now describe this method, on a boids application.

**THE METHOD**

**Relationship between *n*-trees and organizations**

As said above, the hierarchical subdivision of space can be represented by a tree whose root is the whole considered space and branches and leaves are subdivisions of this space. The space is subdivided so that there is a constant, predefined, maximum number of entities (which, per analogy with the N-body problem, we will call *bodies)* in each subspace.

We call *cell* a node of the tree and its associated subspace. Depending on the way the space is decomposed, the number of sub cells varies. Often 2D spaces are cut in four cells yielding *4*-trees or "quadtrees", and 3D spaces are decomposed in eight cells, producing *8*-trees or "octrees".

The number of bodies per leaf cell is variable. Some methods can put only one body per cell, others put several bodies in one cell. Bodies are only "present" in the leaf spaces, and super cells aggregate the data of their sub cells.

It is necessary to subdivide a cell when the maximum number of bodies per cell is reached, and *n* cells merge into one when there is less than the maximum number of bodies per cell in all the *n* cells. Therefore, this subdivision is adaptive, and changes during time.

We call a "level" all the cells of the tree at a given depth in the tree. The nature of the subdivision yields many subdivided cells where there are many bodies. The intuition of the method is that it is easy to identify dense groups of bodies (that is places where interactions are stronger) by looking at a representation of the subdivided space (figure 1).

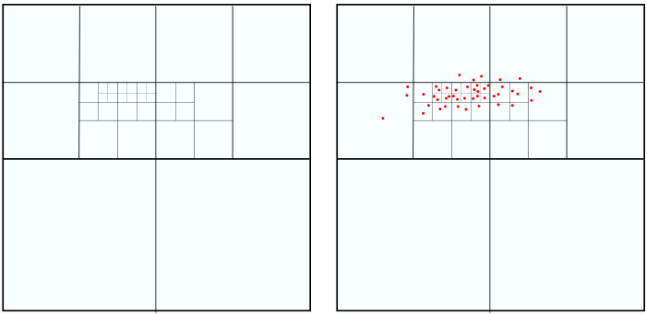

Figure 1 : The space subdivision, with and without bodies. Each cell contains a maximum of 3 bodies.

**Finding organizations with the n-tree**

Our method relies on the use of the *n*-tree to detect organizations, therefore sharing the use of this tree both for the FMM and for the organization detection.

The main idea is to consider only the lower levels of the tree as shown on figures 2 and 3, and then to cut this set of cells in groups of adjacent cells. The tree in figure 3 represents the subdivided space of the figure 2. There are at most three bodies per cell here. The search for organizations begins by considering only the lower levels of the tree. That is, cutting the bottom of the tree (here under the horizontal dotted line). In this restricted set of cells, we remove all cells that do not contain bodies (in grey on the figure). Notice that this selects only "leaf" cells. This set must then be cut in groups of adjacent cells. These groups can be considered a good approximation of searched organizations.

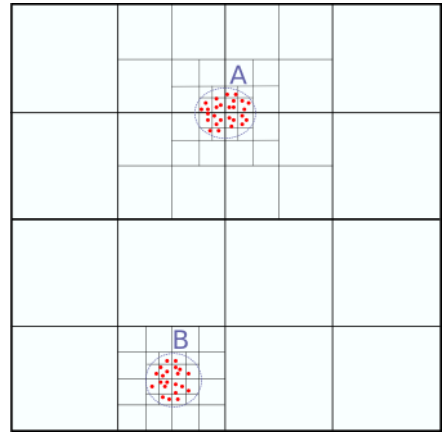

Figure 2 : Two organizations A and B and the subdivided space

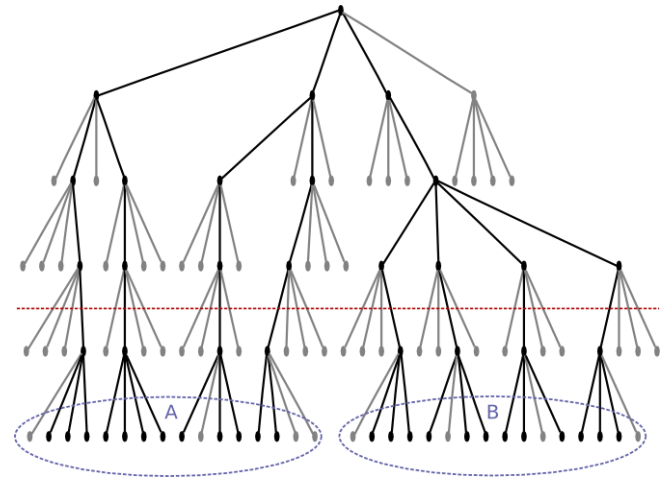

Figure 3 : The A and B organizations in the n-tree

In two dimensions, one can also see the subdivisions of space as being higher in a third dimension. The more space is subdivided, the higher a point in space is. As subdivision matches body density, the denser the bodies are the higher is their third coordinate. The idea is then to cut this representation and consider only the cut parts to search for dense groups of bodies, as shown on Figure 4.

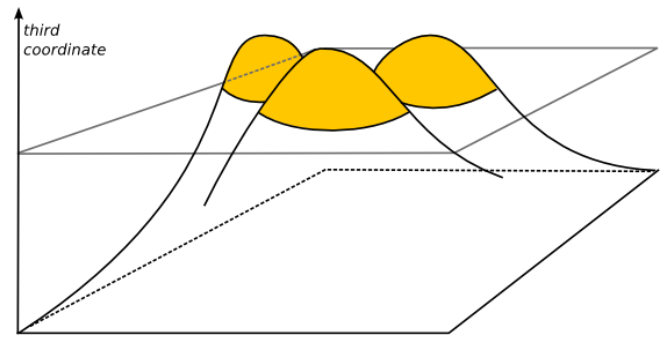

Figure 4 : Conceptual representation of the method.

**Algorithm**

The organization detection algorithm does not work directly on the n-tree, but on a set of cells C corresponding to some levels of the n-tree. They correspond to all the cells of the tree that are deeper than a given depth d.

In this set, we keep only the leaf cells, as only these cells "contain" bodies.

The algorithm will then try to cut this set in groups of adjacent cells. We will consider these groups as organizations. Adjacent cells are cells that share a face, an edge or a point (Figure 5).

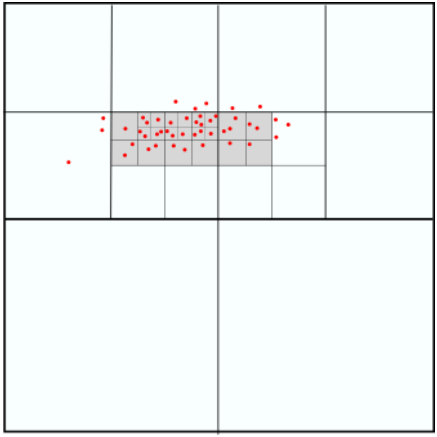

Figure 5 : Adjacent cells delimit organizations

The creation of the groups can be done easily with a $O(n^2)$ algorithm, with $n$ the number of cells in C. Consider one cell $c$ taken at random in C and remove it from C. Consider all the other cells of $C$ and check if they share a face, edge or point with $c$. If they do, put them in the same group and remove them from C.

When there are no more cells of C adjacent to $c$, and if C still contains cells, create another group, choose randomly another cell $c'$ and restart the procedure. Do this until C is empty.

```
Algorithm GroupCells( C ) returns G
C: set of cells
G: set of sets of cells
c: cell
g: set of cells
While not C.isEmpty do
    g <- new group
    c <- C.removeRandom
    G.add( g )
    g.add( c )
    ForAll cell d in C do
        If d.shareFaceOrEdgeOrPoint( g )
        then
            C.remove( d )
            g.add( d )
        EndIf
    EndForAll
EndWhile
```

However, the group creation can be achieved faster by using the *n*-tree. It is not necessary to explore the entire set C to find adjacent cells. Instead, it is possible to explore the tree.

To find the adjacent cells of a cell c taken at random in C and removed from it, start from the root cell and look for all child cells that could intersect or contain an area around the cell c. Do this recursively until you reach leaf cells. Only in this restricted set of cells, search for adjacent cells. Restart this procedure for each newly found cell until no more adjacent cell is found.

```
Algorithm GroupCells2( C ) returns G
C: set of cells
G: set of sets of cells
c, n: cell
g, h: set or cells
While not C.isEmpty do
    g <- new group
    h <- new group
    c <- C.removeRandom
    G.add(g)
    g.add(c)
    h.add( neighborsOf( tree.getRoot, c, C, area ) )
    While not h.isEmpty do
        ForAll cell d in h do
            If d.shareFaceOrEdgeOrPoint( g )
            Then
                C.remove( d )
                h.remove( d )
                g.add( d )
                h.add( neighborsOf( tree.getRoot, d, C, area ) )
            Else
                h.remove( d )
            EndIf
        EndForAll
    EndWhile
EndWhile

Method NeighborsOf( n, c, C, tree, area ) returns h
c, n: cell
C, h: set of cells
tree: n-tree
area: float
If not n.isLeaf then
    ForAll cell d in n.subCells do
        If n.containsOrOverlap( area, c ) then
            h.add( NeighborsOf( d, c, C, area ) )
        EndIf
    EndForAll
Else
    If C.contains( n ) then
        h.add( n )
    EndIf
EndIf
```

The difficulty in this method is that it is difficult to compute which cells are intersecting with the area around a leaf cell. A simple approximation can be to use a cube or square instead of a sphere or disk around the considered cell.

**RESULTS**

The results presented here are in two dimensions. This allows an easy representation of the levels. On each figure, a boid simulation is shown on the left and the space subdivision is shown on the right.

The boid simulation uses several species that repulse one another and form groups of high interaction. The space subdivision is in two dimensions, with a third dimension showing the depth in the quadtree. Levels close to the root are in darker and near the leaves are lighter. The cut threshold has been set to 5 levels, and levels higher than this have been colored in red, they show the organizations detected. The maximum number of bodies per cell is ten.

The Figure 6 shows the boid simulation at different time steps.

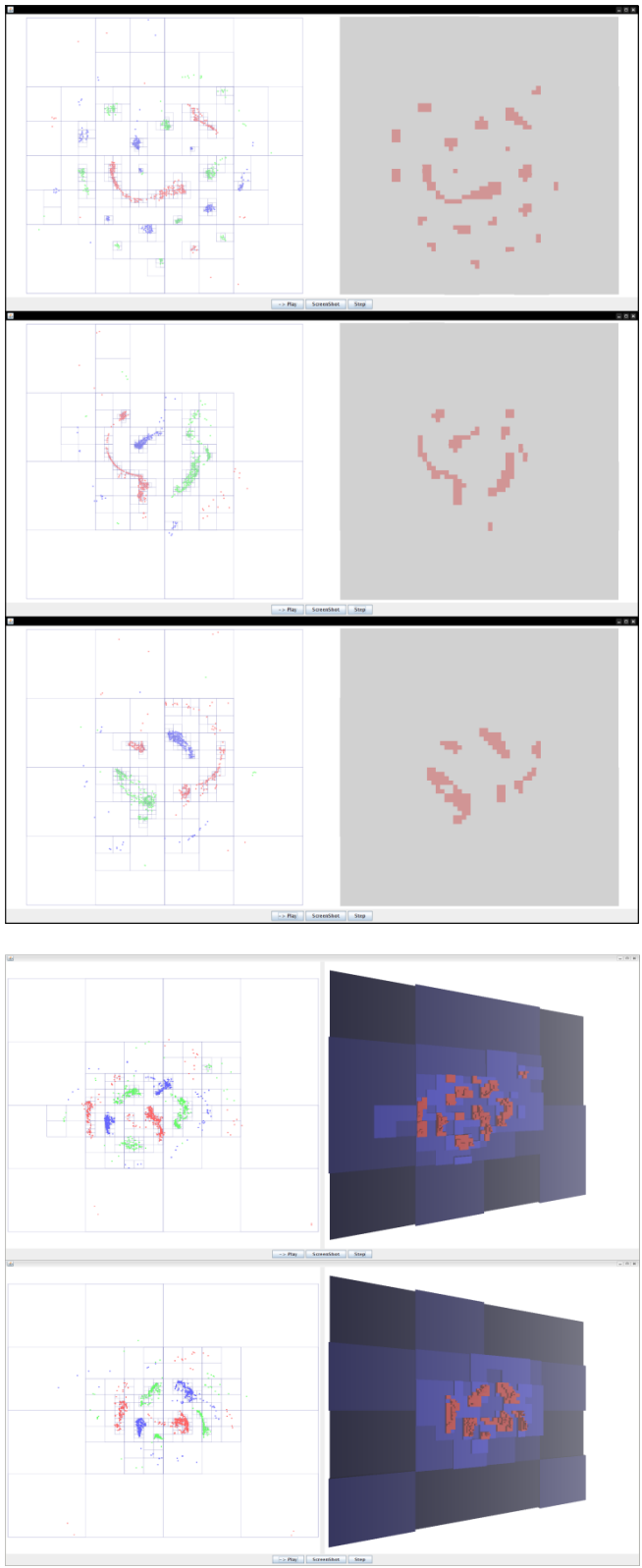

Figure 6 : Left, the boids in different colors. Right, the organizations detected, in red[1]

---

[1] The last two screenshots of Figure 6 depict the same results as the first three. Some readers have preferred this representation, others the flat one.

# FUTURE WORK

The organizations detected are well matched to what a human observer perceives. Nonetheless, evidence that is more objective must be used to validate their interest. For that, we will use two methods.

For the first method we will need to transform our set of boids into a complete weighted graph, where the vertices will be the boids, the edges the interaction between the boids, and the weight of an edge the distance between the boids corresponding to the vertices of this edge. The modularity of the detected organization can then be computed.

The second method is computing the fuzzy kappa index of the simulation spaces with and without the organizations [Hagen-Zanker 2006]. This method was invented to compare actual maps to ones generated by cellular automata methods. It should be no problem to use it here.

These two methods are currently being implemented.

# CONCLUSION

We presented an organization detection method based on the reuse of the n-tree of fast multipole methods. Such methods are used in a wide range of problems. It offers a fast algorithm that uses the calculations of another one to speed up its computation. The detected groups of bodies are close approximation of the real organizations.

# REFERENCES

Beatson R. and Greengard L. *A short course on fast multipole methods*, http://math.nyu.edu/faculty/greengar/. Link is alive on 01/18/08.

Coulomb C.-A. 1785, *Premier Mémoire sur l'Électricité et le Magnétisme,* Paris

Dutot A., 2005, *Distribution dynamique adaptative à l'aide de mécanismes d'intelligence collective*, PhD thesis, Le Havre University

Frontier S., Pichot-Viale D.., *Écosystèmes, Dunod, 1998.*

Hagen-Zanker A., 2006, *Map comparison methods that simultaneously address overlap and structure*, Journal of Geographical Systems, Springer, vol. 8(2), pages 165-185, July.

Leonard A. 1980, « Vortex methods for flow simulation », *Journal of Computational Physics*, vol. 37, 1980, p. 289-335.

Newton I., 1686 *Philosophiae Naturalis Principia Mathematica*, London.

Reynolds, C. W., 1987. "Flocks, Herds, And Schools: A Distributed Behavioural Model". *Computer Graphics*. Vol. 21, No.4.

Simon H. 1996, *The Sciences of the Artificial (3rd Edition)* MIT Press

Tranouez Pierrick 2005, *Penicillo haere, nam scalas aufero*, PhD thesis, Le Havre University